\documentclass[runningheads]{llncs}
\usepackage[T1]{fontenc}
\usepackage{graphicx}
\usepackage{amsmath}
\usepackage{amsfonts}
\usepackage{booktabs}
\usepackage{pifont}
\usepackage{algorithm,algpseudocode}

\DeclareMathOperator*{\argmin}{arg\,min}
\newcommand{\xmark}{\ding{55}}
\newcommand{\vmark}{\ding{51}}
\newcommand*{\Break}{\textbf{break}}

\begin{document}
\title{Planar-SfM: Camera Pose Estimation via Homography Graph Embeddings}
%
%\titlerunning{Abbreviated paper title}
% If the paper title is too long for the running head, you can set
% an abbreviated paper title here
%

\author{Gabi Pragier$^{*}$ \and  
Matan Karklinsky$^{*}$ \and 
David Ungarish$^{*}$ \and 
Avi Ben-Cohen}
% \author{Gabi Pragier \and  Matan Karklinsky \and David Ungarish \and Avi Ben-Cohen}
%
\authorrunning{G. Pragier et al.}
% First names are abbreviated in the running head.
% If there are more than two authors, 'et al.' is used.
%
\institute{Amazon Prime Video Live Events\\ 
\email{\{gabipragier,davidunga\}@gmail.com} \, \email{\{mtn,avibc\}@amazon.com}\\
$^{*}$Equal contribution}

% \institute{Amazon Prime Video Live Events\\ \email{\{gpragier, mtn, uda, avibc\}@amazon.com}}
%
\maketitle              % typeset the header of the contribution
\begin{abstract}
Structure from Motion (SfM) systems traditionally struggle with planar scenes, where standard epipolar geometry-based methods become degenerate. Rather than viewing planar surfaces as a limitation, we propose a unified framework that leverages them as a source of geometric constraints. Our key insight is that each planar surface visible across multiple views provides an independent estimate of relative camera poses through homography decomposition. By aggregating estimates from multiple planes or even from a single dominant plane we achieve robust pose recovery in scenarios where traditional methods fail.
We introduce a novel graph-based approach that constructs a pose-graph from homography estimates and employs spectral embedding to identify and filter unreliable edges. Our method maps homography-based pose estimates onto the real line based on their geometric and visual consistency, enabling efficient extraction of a maximally consistent spanning tree for pose recovery. This approach naturally handles both highly planar scenes, such as indoor sports arenas, and general $3$D environments.
We demonstrate superior performance on basketball court imagery where existing methods struggle, while matching or exceeding state-of-the-art results on unconstrained outdoor scenes from the IMC Phototourism benchmark.

\end{abstract}

\section{Introduction}
\label{sec:intro}

Structure from Motion (SfM) is fundamental to computer vision, enabling $3$D scene reconstruction from multiple viewpoints~\cite{lindenberger2021pixel,ozyesil2017survey,schoenberger2016sfm,snavely2006photo}. Its applications span augmented reality~\cite{marchand2016pose}, $3$D reconstruction~\cite{choy2016unified,rezende2016unsupervised,schonberger2016pixelwise}, SLAM~\cite{campos2021orb,mur2015orb}, and modern neural rendering techniques~\cite{kerbl3Dgaussians,mildenhall2020nerf}.
A significant limitation of existing SfM systems is their difficulty handling scenes that are dominated by a single planar surface. In such cases, the standard approach of estimating camera poses through epipolar geometry is subject to a two-parameter ambiguity~\cite{hartley2003multiple}, resulting in a multifold solution space. Consequently, the estimation of the epipolar geometry becomes ill-posed and is prone to arbitrary determination by outlier correspondence. This is particularly problematic in many real-world applications, such as sports broadcasting in indoor arenas or architectural documentation, where large planar surfaces are common.

While planar surfaces are often viewed as a challenge for SfM, we argue they present an opportunity. Even in general scenes, decomposing the environment into multiple planar segments offers distinct advantages. Each planar surface provides an independent estimate of relative camera poses through homography decomposition, and observing multiple planes offers complementary constraints that enable more robust pose recovery. By leveraging multiple such independent estimates, we can achieve more robust camera pose recovery than methods relying on point-based epipolar geometry alone.
Our key insight is that by treating scenes as collections of planar surfaces, whether they actually contain one plane, multiple planes, or general $3$D structure, we can develop a unified approach that handles all cases effectively. We achieve this through three main contributions:
\begin{itemize}
    \item An embedding framework that filters unreliable homography-based pose estimates, enabling camera motion recovery even in challenging single-plane scenarios where traditional methods struggle.
    \item A multi-plane aware pose estimation strategy that leverages independent estimates from different planar regions to enhance accuracy, even in general $3$D scenes.
    \item Comprehensive evaluation demonstrating superior performance both on planar scenes where traditional methods struggle, and on general scenes where our approach matches or exceeds state-of-the-art results.
\end{itemize}

\section{Related Work}
\label{sec:related_work}

Conventional SfM pipelines typically follow a multi-stage process: feature detection across images~\cite{detone2018superpoint,dusmanu2019d2,lowe2004distinctive} followed by feature matching across pairs~\cite{brachmann2019neural,lindenberger2023lightglue,sarlin2020superglue,sun2021loftr}, and then estimation of relative camera poses (two-view geometry). Additional views are then incrementally registered in a sequential manner, which is prone for drift and error accumulation. A global optimization step, typically bundle adjustment~\cite{triggs2000bundle}, refines camera parameters and $3$D point coordinates jointly. However, due to the non-convexity of this optimization, the accuracy of the final solution strongly depends on the quality of the initial estimates. To better handle structured environments, several methods have incorporated prior geometric knowledge. For example,~\cite{fischler1981random} proposed a RANSAC-based scheme that exploits known planar structures by operating over camera trajectories. Similarly,~\cite{pollefeys2002surviving} introduced a mechanism for filtering out degenerate images based on scene priors. Nonetheless, these sequential pipelines remain prone to local minima and often fail to reach globally optimal solutions. Recently, learning-based approaches have formulated SfM as a differentiable process~\cite{teed2020deepv2d,zhou2020deeptam}, achieving state-of-the-art performance. Despite this progress, these methods lack explicit enforcement of multi-view geometric constraints, which can limit their accuracy in planar scenes.

\section{Method}
\label{sec:method}
\subsection{Setting}
\label{subec:setting}
The input consists of a set of $n$ images capturing a scene containing one or more planar surfaces, where the number of surfaces is unknown. The underlying cameras are assumed to be calibrated, that is with known intrinsic matrices which specify the focal length and principal point parameters. The goal is to recover the camera poses $(R_i, c_i) \in SE(3), \  i=1,\ldots,n$, where $R_i$ and $c_i$ represent the orientation and position, respectively, of the $i$-th camera. We denote the relative pose between cameras $i$ and $j$ by
$(R_{ij} , c_{ij}) \in SE(3)$, where $R_{ij} = R_j R_i^T$ is their relative
orientation and $c_{ij} \propto R_j(c_i - c_j)$ is their relative position. In addition, we denote by $n_\pi \in \mathbb{S}^2$ the unit-length normal vector to the planar surface $\pi$.

\begin{figure*}[t!]
  \centering
  \includegraphics[width = \textwidth]{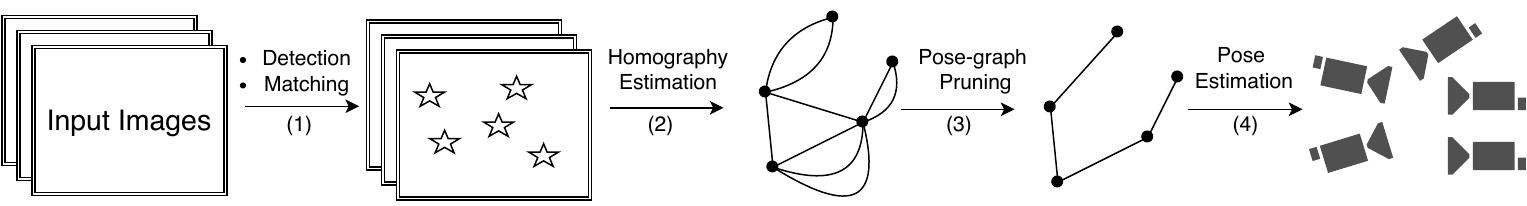}
  \caption{\textbf{Method Overview:} Our pipeline processes input images through four main stages: (1) feature detection and matching, (2) homography-based pose-graph construction, (3) graph pruning via real-line embedding, and (4) camera pose recovery.}
  \label{fig:pipeline}
\end{figure*}

\subsection{Overview}
\label{subsec:overview}
Our method processes the input images in several stages, as illustrated in Figure~\ref{fig:pipeline}. First, we construct a pose-graph where edges represent homographies estimated between image pairs (Section~\ref{subec:homography pose-graph}). Each homography, when arising from a genuine planar surface, provides both the relative pose between the corresponding cameras and the surface normal through decomposition.

To handle potentially erroneous homographies, we develop a two-stage filtering approach. We first compute similarity scores between homographies based on their geometric and visual consistency (Section~\ref{subsec:edge_similarity_score}). These scores incorporate both image-space evidence and structural validation against essential matrix-based estimates. We then use these scores to embed the multi-edges of our pose-graph onto the real line, allowing us to identify and retain only the most consistent set of edges (Section~\ref{subsec:embedding}).

Finally, we recover camera poses using the filtered pose-graph (Section~\ref{subsec:pose_estimation}). We first consolidate multiple homography-based estimates within each retained edge into a single relative pose estimate. We then recover camera orientations by propagating through a spanning tree, and positions by extending to a bi-connected graph and solving a global optimization problem.

\subsection{Homography Pose-Graph}
\label{subec:homography pose-graph}
Based on matched feature points between image pairs, we estimate $k_{ij} \geq 0$ homographies $\{H_{ij}^q\}_{q=1}^{q=k_{ij}}$ for each image pair $(i,j)$, where each homography corresponds to a plane visible in both images. We represent the scene as a pose-graph $G=(V, E)$, an undirected multigraph where each node~$i \in V$ represents a camera pose to be determined, and we connect nodes $i$ and $j$ with $k_{ij}$ edges whenever we find homographies between their images.

A key property of homographies is that, given calibrated cameras, each homography $H_{ij}$ between images $i$ and $j$ of a plane $\pi$ can be decomposed~\cite{ma2004invitation} into
\begin{equation}\label{eq:decomposition}
H_{ij} \mapsto (R_{ij}, c_{ij}, n_{ij}),
\end{equation}
where $n_{ij} = R_i n_{\pi}$, and with $(R_{ij}, c_{ij})$ being precisely the relative pose defined in Section~\ref{subec:setting} and is invariant to $\pi$. Thus, when $H_{ij}^q$ arises from a genuine planar surface (rather than erroneous feature correspondences), its decomposition provides both the underlying relative pose and the plane normal expressed in camera $i$'s coordinate system. We next describe how to account for homographies arising from incorrect correspondences.

\subsection{Edge Similarity Score}
\label{subsec:edge_similarity_score}
While each homography decomposition (Eq.~\ref{eq:decomposition}) provides an estimate of relative camera pose, not all decompositions are reliable. However, since a spanning tree of the pose-graph $G$ suffices to recover camera poses, we can afford to be selective and retain only the most reliable edges. To identify these, we develop an embedding-based approach that maps each multi-edge onto the real line, allowing us to utilize spanning tree algorithms while accounting for geometric and visual 
consistency.

The key insight is that reliable homographies induced by the same physical surface should exhibit consistent geometric and visual properties. We capture this consistency through a similarity matrix $S \in \mathbb{R}^{|E| \times |E|}$, where each entry $S_{ij,lm}$ represents the expected consistency between incident multi-edges $e_{ij}, e_{lm} \in E$:
\begin{equation}\label{eq:S}
S_{ij,lm} = \sum_{q=1}^{k_{ij}} \sum_{q'=1}^{k_{lm}} P^{q,q'}_{ij,lm} \eta(n^q_{ij}, n^{q'}_{lm}).
\end{equation}
Here, $\eta(n_a, n_b) = \exp\left(-\arccos^2\left({n_a^T n_b}\right)\right)$ measures the angular similarity between surface normals, and $P^{q,q'}_{ij,lm}$ represents the probability ($\sum_{q, q'} P^{q,q'}_{ij,lm} = 1$) that homographies $H^q_{ij}$ and $H^{q'}_{lm}$ are not erroneous and correspond to the same physical surface. This probability is computed by combining image-space evidence and structural validation. Specifically, for any two homographies $H^q_{ij}$ and $H^{q'}_{lm}$ we define:
\begin{equation}
P^{q,q'}_{ij,lm} \propto \mathcal{O}(H^q_{ij}, H^{q'}_{lm}) w_{ij}^q w_{lm}^{q'},
\end{equation}
where $\mathcal{O}(H^q_{ij}, H^{q'}_{lm})$ which counts the number of feature points that are inliers in both homographies, provides image-space evidence for whether the two homographies correspond to the same physical surface. The confidence scores $w_{ij}^q$ provide structural evidence for the validity of each homography by measuring its agreement with an independently estimated orientation. Specifically, these scores are defined using the angular similarity between the relative orientation $R_{ij}^q$ obtained from the homography decomposition (Eq.~\ref{eq:decomposition}) and a plane-independent auxiliary orientation~$R_{ij}^{e}$ estimated from the decomposition of the essential-matrix~\cite{hartley2003multiple} between images $i$ and $j$. Specifically, 
\begin{equation}\label{eq:w_ij}
w_{ij}^q \propto \exp\left(-\epsilon^2\left(R_{ij}^q, R_{ij}^{e}\right)\right),    
\end{equation}
where
$$
\epsilon(R_a, R_b) = \arccos\left(\frac{\text{trace}(R_a^T R_b) - 1}{2}\right)
$$
is the geodesic distance on $SO(3)$. 

\subsection{Pose-Graph Embedding and Pruning}
\label{subsec:embedding}
To recover camera poses, we need a spanning tree of the pose-graph $G$. However, not all edges are equally reliable since some homographies may arise from incorrect correspondences or degenerate configurations. Our goal is to identify a spanning tree whose edges are mutually consistent, corresponding to valid homographies from compatible planar surfaces. To achieve this, we first embed the multi-edges onto the real line (using the similarity matrix $S$ from Eq.~\ref{eq:S}) by solving 
\begin{equation}\label{eq:embedding}
\begin{aligned}
h = &\argmin_{\tilde{h} \in \mathbb{R}^{|E|}} \quad  \sum_{ij, lm \in E} \left(\tilde{h}_{ij} - \tilde{h}_{lm}\right)^2 S_{ij, lm} \\
&\textrm{subject to} \quad  \tilde{h}^T D \tilde{h} = 1,
\end{aligned}
\end{equation}
where $D$ is the diagonal matrix of row sums of $S$. This formulation encourages similar edges (those with high $S_{ij,lm}$) to be embedded close together, while the constraint ensures a non-trivial solution that accounts for edge importance through $D$. The optimal solution is given by the eigenvector associated with the second-smallest eigenvalue of $I - D^{-1}S$~\cite{belkin2003laplacian}, as the eigenvector of the smallest eigenvalue, namely the constant vector, is uninformative for our purpose.

With the embedding $h$ established, and since its scale is arbitrary, we define an inconsistency measure for any spanning tree $\tilde{T}$ of the pose-graph $G$ as the magnitude of the support of its embedding $\max_{ij,lm \in E(\tilde{T})} |h_{ij} - h_{lm}|$. The optimal spanning tree is then obtained by solving:
\begin{equation}\label{eq:IMOptimProblem}
    T = \argmin_{\tilde{T} \subseteq G} \left\{ \max_{ij,lm \in E(\tilde{T})} |h_{ij} - h_{lm}| \right\},
\end{equation}
whose solution $T$ is given by the minimum-range-spanning-tree of the weighted graph $G$ whose weights are given by the embedding $h$ obtained from Eq.~\ref{eq:embedding}. The step-by-step process for determining the minimum-range-spanning-tree $T$ is outlined in Algorithm~\ref{alg:MDST}, provided in Appendix~\ref{sec:minimal range spanning tree}.

\subsection{Pose Estimation}
\label{subsec:pose_estimation}

Given the minimal-range spanning tree $T$, we next describe how to recover the camera poses. A key advantage of our approach is that any relative pose $(R_{ij}, c_{ij})$ is invariant to the underlying plane (as noted in Eq.~\ref{eq:decomposition}). This invariance enables us to leverage multiple planar regions within each multi-edge to enhance accuracy: when multiple homographies are estimated between the same image pair, whether from distinct physical planes in a general $3$D scene or from different segmentations of a single dominant plane, their decompositions provide independent estimates of the same underlying relative pose. By aggregating these estimates, we effectively reduce noise and improve robustness.

Specifically, we consolidate the multiple pose estimates within each edge $e_{ij}$ in $T$ into a single estimate~$(\hat{R}_{ij}, \hat{c}_{ij})$ by projecting their confidence-weighted averages (weights defined in Eq.~\ref{eq:w_ij}) onto $SO(3)$ and $\mathbb{S}^2$, respectively:
\begin{equation}
\hat{R}_{ij} = \Pi_{SO(3)}\left(\sum_q w_{ij}^q R_{ij}^q\right), \quad 
\hat{c}_{ij} = \Pi_{\mathbb{S}^2}\left(\sum_q w_{ij}^q c_{ij}^q\right).
\end{equation}
Camera orientations $\hat{R}_i$ are then recovered by setting one camera's orientation to the identity matrix and propagating through $T$ using the estimates $\hat{R}_{ij}$. 
For the task of recovering camera positions from relative positions to be well-posed, the underlying graph must exhibit parallel rigidity, a property we empirically found to be equivalent in many cases to the graph being bi-connected (meaning that if any one node were to be removed the graph will still remain connected). As no tree is bi-connected, the spanning tree $T$ is inadequate for determining the camera positions. We therefore extend $T$ to a bi-connected graph by iteratively adding edges that minimize embedding inconsistency. Finally, using the recovered orientations, we transform each relative position as $\hat{c}'_{ij} = \hat{R}^T_j \hat{c}_{ij}$ to obtain $\hat{c}'_{ij} \propto c_i - c_j$, from which absolute positions are recovered using least-unsquared-deviations~\cite{ozyesil2015stable}.

\section{Experiments}

\subsection{Datasets and Setup}
\begin{figure}
    \centering
    \includegraphics[width=\columnwidth]{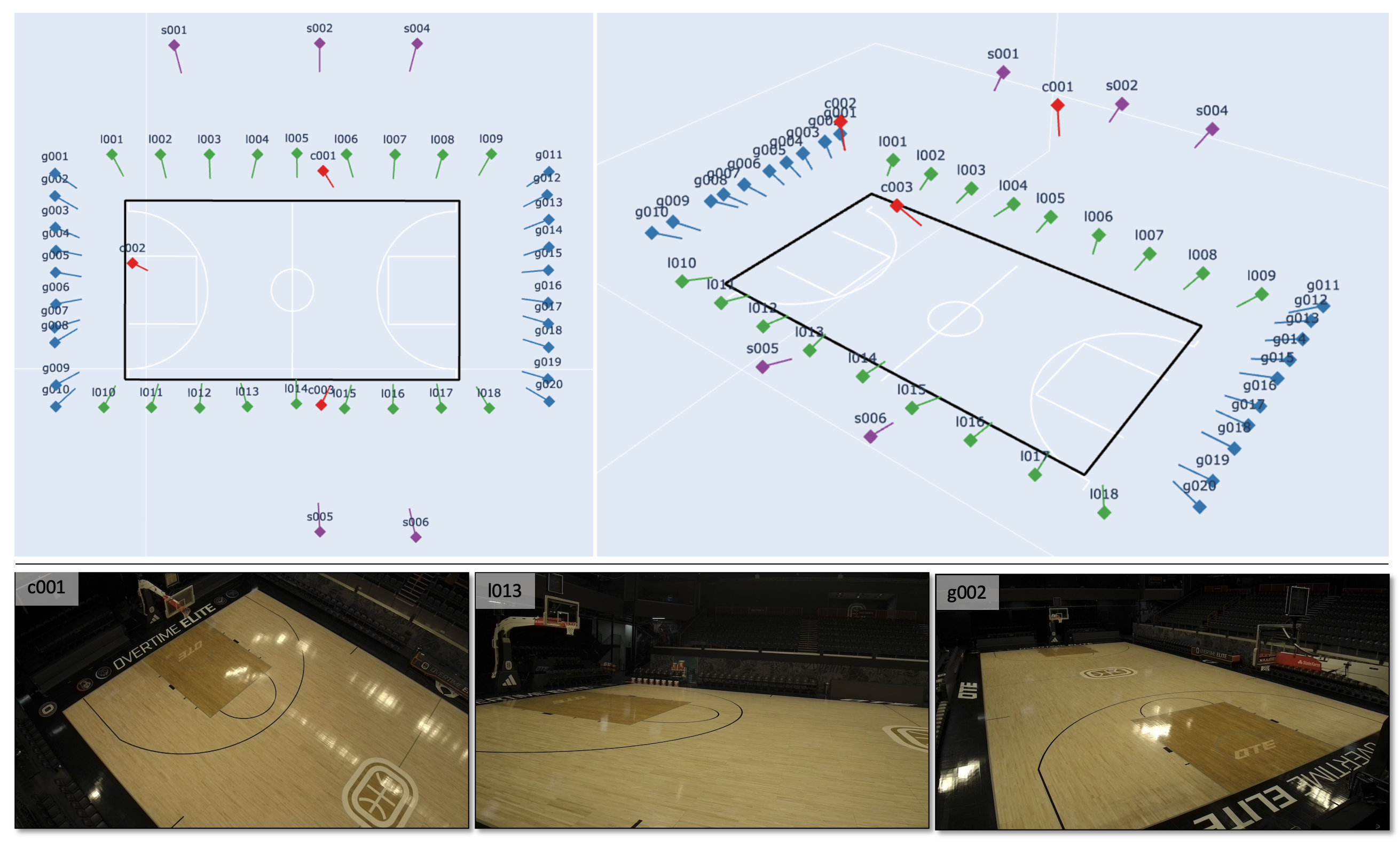}
    \caption{\textbf{OTE Dataset}. Top: Camera placements and poses visualized from two views. Bottom: Example images, captured by three cameras.}
    \label{fig:ote}
\end{figure}

We evaluated our method on two datasets (\textrm{i}) The Overtime Elite (OTE) basketball arena dataset consisting of $46$ calibrated cameras positioned around a basketball court (Figure~\ref{fig:ote}), representing a real-world sports broadcasting scenario, and (\textrm{ii}) the IMC Phototourism dataset~\cite{Jin2020} with $8$ outdoor scenes captured under unconstrained conditions. We used SuperPoint~\cite{detone2018superpoint} for feature extraction, and SuperGlue~\cite{sarlin2020superglue} for feature matching. For OTE, we matched each image with its two nearest neighbors from each side, exploiting the ordered camera arrangement. For IMC, we performed exhaustive pairwise matching.
We compared against GLOMAP~\cite{pan2024GLOMAP} for OTE, and against both classical methods (COLMAP, PixSfM~\cite{schoenberger2016sfm,lindenberger2021pixel}) and learning-based methods (VGGSfM, DFSfM~\cite{wang2024vggsfm,he2024detector}) for IMC. Following~\cite{lindenberger2021pixel,sarlin2020superglue}, we report the area under curve~(AUC) of relative orientation errors.

\subsection{Results}
For OTE, our method outperforms GLOMAP at $3^\circ$ and $5^\circ$ thresholds while remaining competitive at $1^\circ$, demonstrating the advantage of our planar-aware approach on basketball court scenes, see Table~\ref{tab:ote_results}. On IMC, we achieve state-of-the-art performance across all thresholds, see Table~\ref{tab:imc_results}, showing that our method is capable at generalizing well to unconstrained, non-planar scenes.

\begin{table}
\centering
\caption{Camera pose estimation results on the OTE dataset. Our method achieves better accuracy than GLOMAP at the $3^\circ$ and $5^\circ$ thresholds, with comparable performance at $1^\circ$.}
\begin{tabular}{lccc}
\hline
\textbf{Method} & \textbf{AUC@1$^\circ$} & \textbf{AUC@3$^\circ$} & \textbf{AUC@5$^\circ$} \\
\hline
GLOMAP \cite{pan2024GLOMAP} & \textbf{67.45} & 84.72 & 91.12 \\
Ours & 67.39 & \textbf{86.55} & \textbf{95.55} \\
\hline
\end{tabular}
\label{tab:ote_results}
\end{table}

\begin{table}
    \centering
    \setlength{\tabcolsep}{4pt}  % default is 6pt
    \caption{Camera pose estimation results on the IMC dataset~\cite{Jin2020}. Table reproduced from \cite{wang2024vggsfm}, with the last row ("Ours") added.}\label{tab:imc_results}
    \begin{tabular}{lccc}
    \toprule
    \textbf{Method} & \textbf{AUC@3$^\circ$} & \textbf{AUC@5$^\circ$} & \textbf{AUC@10$^\circ$} \\
    \midrule
    DeepSFM               & 10.27 & 19.36 & 31.35 \\
    PoseDiffusion         & 12.31 & 23.17 & 36.82 \\
    \midrule
    COLMAP (SIFT+NN)      & 23.58 & 32.66 & 44.79 \\
    PixSfM (SIFT + NN)    & 25.54 & 34.80 & 46.73 \\
    PixSfM (LoFTR)        & 44.06 & 56.16 & 69.61 \\
    PixSfM (SP + SG)      & 45.19 & 57.22 & 70.47 \\
    DFSfM (LoFTR)         & 46.55 & 58.74 & 72.19 \\
    \midrule
    VGGSfM w/o Joint      & 38.23 & 51.60 & 68.35 \\
    VGGSfM                & 45.23 & 58.89 & 73.92 \\
    Ours                  & \textbf{49.11} & \textbf{60.32} & \textbf{76.10} \\
    \bottomrule
    \end{tabular}
\end{table}

\subsection{Ablation Study}
\label{sec:Ablation}
We evaluated the impact of our method's two key components confidence scoring~(Section~\ref{subsec:edge_similarity_score}) and graph pruning~(Section~\ref{subsec:embedding}) on the OTE dataset. To disable confidence scoring we assigned random weights to homographies, and to disable pruning we replaced the minimum range spanning tree with a random spanning tree. We averaged the results over $100$ random seeds. 

The results, shown in Table~\ref{tab:ablation}, demonstrate that both components are crucial and complementary. Without pruning~(second row in Table~\ref{tab:ablation}), the underlying graph structure remains unreliable regardless of the weights used. Without proper confidence scores~(third row in Table~\ref{tab:ablation}), pruning alone provides some benefit but falls short of the full method. The dramatic performance drop when either component is disabled highlights their effectiveness in filtering outliers and reinforcing reliable estimates.

\begin{table}
\caption{Ablation study results on the OTE dataset. Disabling either component leads to significant accuracy drops, demonstrating their complementary roles in outlier rejection.}
\label{tab:ablation}
\centering
\setlength{\tabcolsep}{4pt}  % default is 6pt
\begin{tabular}{ccccc}
\toprule
\textbf{Pruning} & \textbf{Conf. Scores} & \textbf{AUC@1$^\circ$} & \textbf{AUC@3$^\circ$} & \textbf{AUC@5$^\circ$} \\
\midrule
\vmark & \vmark & 67.39  & 86.55  & 95.55 \\
\xmark & \vmark &  3.303 & 11.566 & 17.909 \\
\vmark & \xmark     &  0.808 &  3.636 &  6.263 \\
\xmark & \xmark     &  0.434 &  2.646 &  4.910 \\
\bottomrule
\end{tabular}
\end{table}

\section{Summary}
We presented Planar-SfM, a Structure from Motion framework that leverages planar surfaces as geometric constraints rather than viewing them as a limitation. We constructed a pose-graph from homography estimates between image pairs, where each homography provides an independent estimate of relative camera pose through decomposition. To filter unreliable estimates, we applied a spectral embedding approach that maps edges onto the real line based on geometric and visual consistency, enabling extraction of a maximally consistent spanning tree for pose recovery. We demonstrated superior performance on basketball court imagery where traditional epipolar geometry-based methods struggle, while matching or exceeding state-of-the-art results on the IMC Phototourism benchmark for general outdoor scenes.

%\subsubsection{Acknowledgements} xxxx

%
% ---- Bibliography ----
%
% BibTeX users should specify bibliography style 'splncs04'.
% References will then be sorted and formatted in the correct style.
%
\bibliographystyle{splncs04}
\bibliography{bibliography}

\section*{Appendix}
\appendix

\section{Finding the Minimum Range Spanning Tree}\label{sec:minimal range spanning tree}

Algorithm~\ref{alg:MDST} below for finding the minimum range spanning tree of a weighted undirected graph is the solution of Eq.~\ref{eq:IMOptimProblem}. The idea behind the algorithm for determining such a tree is that there must be an edge weight that is the smallest among all edge weights in the tree. Thus, by fixing a minimum edge weight threshold and constructing a minimum-spanning-tree~\cite{kruskal1956shortest} that excludes edges with weights below this threshold, we can effectively obtain a minimum-bottleneck-spanning-tree which is a spanning tree in which the heaviest edge is as light as possible. This is because any minimum-spanning-tree is required to contain the edge with the lightest weight in the graph~\cite{prim1957shortest}. The validity of this method stems from the fact that any minimum-spanning-trees is also a minimum-bottleneck-spanning-tree~\cite{camerini1978min}, ensuring that the resulting tree will have the lightest possible maximum edge weight.

\begin{algorithm}
	\caption{Minimum Range Spanning Tree}\label{alg:MDST}
	\begin{algorithmic}[1]
		\State {{\bfseries Input:} undirected weighted graph $G=\left(V, E, W=h\right)$}
            \State {Sort edges in increasing order of weight $h_1,\ldots,h_{|E|}$}
		\For{$i=1,\ldots, |E|-(|V|-1)$}
		% \Comment{any spanning tree contains $n-1$ edges}
            \State{$G_i \gets \left(V, E=\{e_i, \ldots,e_{|E|}\}, W=\{h_i, \ldots,h_{|E|}\}\right)$}
            \If{$G_i$ is not connected}
            \Break
		% \Comment{$h_j \geq h_j, \quad \forall j \geq i$}
            \EndIf
		\State{$T_i \gets \text{minimum-spanning-tree}\left(G_i\right)$}
        \Comment{~\cite{kruskal1956shortest}}
            \State{$d_i \gets \max_{h \in E\left(T_i\right)} \{ h\} - h_i$}
		% \Comment{$h_i = \min_{h \in E\left(T_i\right)} \{ h\}$~\cite{prim1957shortest}}

		\EndFor
            \State{$i^{\ast} \gets \argmin_{i} \{d_i\}$}
		\State {\bfseries Output:} $T_{i^\ast}$
		\Comment{$T_{i^\ast}$ is the solution of Eq.~\ref{eq:IMOptimProblem}}
	\end{algorithmic}
\end{algorithm}

\end{document}